\documentclass[sigconf]{acmart}
\setcopyright{none}
\renewcommand\footnotetextcopyrightpermission[1]{}
\usepackage{graphicx} 
\usepackage{booktabs,siunitx}
\usepackage{amsthm}
\usepackage{multirow, makecell}
\usepackage{subcaption}
\usepackage{float}
\usepackage{dblfloatfix}
\usepackage[english]{babel}
\theoremstyle{plain}
\newtheorem{lemma}[subsection]{Lemma}
\newcolumntype{L}{>{\centering\arraybackslash}m{1.3cm}}

\AtBeginDocument{%
  \providecommand\BibTeX{{%
    \normalfont B\kern-0.5em{\scshape i\kern-0.25em b}\kern-0.8em\TeX}}}

\begin{document}

\title{Variationally Regularized Graph-based Representation Learning for Electronic Health Records}

\author{Weicheng Zhu}
\email{jackzhu@nyu.edu}
\affiliation{\institution{Center for Data Science}
\institution{New York University}
\city{New York}
\state{NY}
\country{USA}}

\author{Narges Razavian}
\email{narges.razavian@nyulangone.org}
\affiliation{\institution{Department of Population Health}
\institution{Department of Radiology}
\institution{NYU School of Medicine}
\city{New York}
\state{NY}
\country{USA}}

\begin{abstract}
  Electronic  Health  Records (EHR) are high-dimensional data with implicit connections among thousands of medical concepts. These connections, for instance, the co-occurrence of diseases and lab-disease correlations can be informative when only a subset of these variables is documented by the clinician. A feasible approach to improving the representation learning of EHR data is to associate relevant medical concepts and utilize these connections. Existing medical ontologies can be the reference for EHR structures, but they place numerous constraints on the data source. Recent progress on graph neural networks (GNN) enables end-to-end learning of topological structures for non-grid or non-sequential data. However, there are problems to be addressed on how to learn the medical graph adaptively and how to understand the effect of medical graph on representation learning. In this paper, we propose a variationally regularized encoder-decoder graph network that achieves more robustness in graph structure learning by regularizing node representations. Our model outperforms the existing graph and non-graph based methods in various EHR predictive tasks based on both public data and real-world clinical data. Besides the improvements in empirical experiment performances, we provide an interpretation of the effect of variational regularization compared to standard graph neural network, using singular value analysis. 
\end{abstract}

\maketitle
\pagestyle{plain}

\section{Introduction}
Electronic Health Records (EHR) are rich sources of information, useful in various predictive tasks in medical application, including mortality prediction, outcomes prediction and phenotyping. The accessibility of the EHR data makes it a feasible resource for scaling screening to large populations. Especially for chronic diseases like Alzheimer's Diseases, early identification before the onset of clinical symptoms can improve the enrollment for clinical trials, and improve effectiveness of the treatments. Previous studies have explored various deep learning methodologies on the EHR \cite{Miotto2016, Cheng2016RiskPW, deepcare}. Learning representations of medical concepts emerges as an important branch \cite{representation, 10.1145/2939672.2939823,DBLP:journals/corr/abs-1810-09593}, and recent research demonstrates the significance of graph structures among medical concepts \cite{DBLP:journals/corr/ChoiBSSS16a, DBLP:journals/corr/abs-1906-04716}. EHR are inherently sparse and structured data with high probability of missing values. Some diseases may be recorded as diagnosis codes, while other existing conditions that are not discussed in the clinical encounter may not be documented. Graph neural network (GNN) has been considered an effective way to generalize Convolutional Neural Networks (CNN) in extracting signals from non-grid structured data \cite{Bruna2013SpectralNA, DBLP:journals/corr/HenaffBL15}. As CNN can focus on the localized features of images or sequences, GNN also enable the model to highlight the significant features and infer the missing features within the topological neighbourhood. Therefore, GNN can be a strong tool for multiple machine learning tasks on EHR, including patient representation learning, medical graph learning and disease prediction.

Our work has the following main contribution on graph-based representation learning of EHR:
\begin{itemize}
    \item We design a novel graph-based model to generalize the ability of learning implicit medical concept structures to a wide range of data source, including short-term ICU data and long-term outpatient clinical data. 
    \item We introduce variational regularization for node representation learning, addressing the insufficiency of self-attention in graph-based models, and difficulties of manually constructing knowledge graph from real-world noisy data sources. The novelty of our work is to enhance the learning of attention weights in GNN via regularization on node representations. With this design, our method outperforms previous graph representation learning method in health predictive tasks based on a clinical EHR and two public EHR datasets.
    \item We provide interpretation on the effect of variational regularization in graph neural networks using singular value analysis, and bridge the connection between singular values and representation clustering.
\end{itemize}
\section{Related Work}
 Among the recent deep learning research on EHR, there are two dominating approaches - extracting temporal signals from time series EHR, and learning embeddings of medical concepts without directly modeling time. \cite{DBLP:journals/corr/ShickelTBR17}. In the first approach, researchers acquire temporal features from sequential biomarkers or encounter data via representation learning methods including recurrent neural networks \cite{DBLP:journals/corr/ChePCSL16, DBLP:journals/corr/ChoiBS15, Diag_LSTM}, convolution blocks \cite{razavian2015temporal, Cheng2016RiskPW} and attention mechanism \cite{song2017attend, DBLP:journals/corr/abs-1907-09538}. The other approach is to train neural networks that express medical concepts with high-dimensional embeddings as learned representations. Med2Vec \cite{DBLP:journals/corr/ChoiBS15} learns to represent EHR codes as embedddings, following the idea of skip-gram \cite{Mikolov2013EfficientEO} in NLP tasks. These previous studies skip an important property of EHR that the diagnosis, labs and procedures are inherently associated with each other. For instance, some diseases co-occur or induce other diseases, and some lab exams indicate certain diseases.

One approach to incorporating the structure of medical concepts into representation learning is to build a medical graph that connects related codes in the EHR. Some previous research models EHR with graph structure: GRAM \cite{DBLP:journals/corr/ChoiBSSS16a} leverages medical ontologies to learn representations of medical concepts. These methods improve representation learning in predictive modeling by incorporating additional structural information or external knowledge. MiME \cite{DBLP:journals/corr/abs-1810-09593} learns hierarchical embeddings of a subset of variables (visits, diagnosis codes, and medications) by building relationships between different levels of hierarchy. These graph-based methods mainly focus on parent-child relationship and rely on the assumption that the EHR data follows hand-designed protocols that may not accurately reflect real world data.

Compared to previous methods, the Graph Convolutional Networks (GCN) \cite{Bruna2013SpectralNA, DBLP:journals/corr/HenaffBL15} are more flexible in learning graph representations. GCN generalizes translation-invariant convolution filters in standard convolutional neural networks (CNN) to a non-Euclidean localized filter \cite{DBLP:journals/corr/HenaffBL15}, that can be applied to various non-grid data. GCN can be applied to learning representations of node features and graph structure through semi-supervised learning for node classification \cite{DBLP:journals/corr/KipfW16}. This work provides an approach for generating labels for unknown nodes given graph structures and node features. Self-attention, comparable to CNN in encoding features from spatial or sequential data \cite{Cordonnier2020On}, requires less computation time and out-preforms CNN in language and vision tasks \cite{NIPS2017_7181, DBLP:journals/corr/abs-1906-05909}. Similar to replacing convolutional block with self-attention, Graph Attention Network (GAT) \cite{velickovic2018graph} attends each node in the graph on its neighbouring nodes and itself to learn localized features instead of using GCN spectral filters. In this architecture, GAT can assign different importance to edges, which increases the model capacity and interpretability, and learns graph structures themselves via attention parameters.

Inspired by self-attention mechanism and GAT, we propose an encoder-decoder GNN by taking each observed EHR code as a node, initially imposing a fully connected graph on them, and implicitly learning their graph structure via self-attention mechanism. A recent work, Graph Convolution Transformer (GCT) \cite{DBLP:journals/corr/abs-1906-04716}, introduces a visit embedding for each patient medical encounter, and combines the visit embedding with the other medical concept embeddings through the Transformer \cite{NIPS2017_7181}. Furthermore, the authors address a problem mentioned in their paper that the Transformer cannot effectively learn the attention parameters from scratch and often leads to uniformly distributed attention weights among medical concepts. GCT in \cite{DBLP:journals/corr/abs-1906-04716} solves the problem by leveraging a pre-defined graph as the guidance of regularization. The graph is constructed by connecting different groups of medical concepts (e.g. diagnosis, labs and procedures) to emulate physicians' decision process.
%

However, the real-world EHR has significant missing data, and the hierarchy among different features cannot be strictly defined. Lab values that correspond to some diseases may not be included in patient's EHR data and vice versa. To design a generalizable model that can work with any groups of variables (diagnosis, labs, procedures, demographics), we do not prune internal connections among diseases that do not match a pre-defined graph, but rather, introduce variational regularization in our encoder-decoder GNN to address the challenges of structure learning without pre-defined graphs. Variational inference has been previously used to improve the representation learning of graph autoencoder \cite{kipf2016variational,NIPS2019_9255} and link prediction with Bernoulli link inference \cite{DBLP:journals/corr/abs-1906-01852,hajiramezanali2019variational}. Previous studies have reported that the regularization effect of KL-term in variational autoencoder \cite{prokhorov-etal-2019-importance} and the strength of regularization is adjustable by scaling on KL-term \cite{DBLP:journals/corr/abs-1901-03416,Higgins2017betaVAELB}. Unlike the autoencoder framework which combines reconstruction loss with the KL-term, here we only use the KL-term to regularize node representations so that the attention weight learning can be enhanced in supervised tasks. We will discuss the benefits of this regularization in section \ref{discussion}.
\section{Methods}
\begin{figure*}
  \centering
  \subcaptionbox{The full architecture of our encoder-decoder model (\textit{Enc-dec})}{
  \includegraphics[width = 0.8\textwidth]{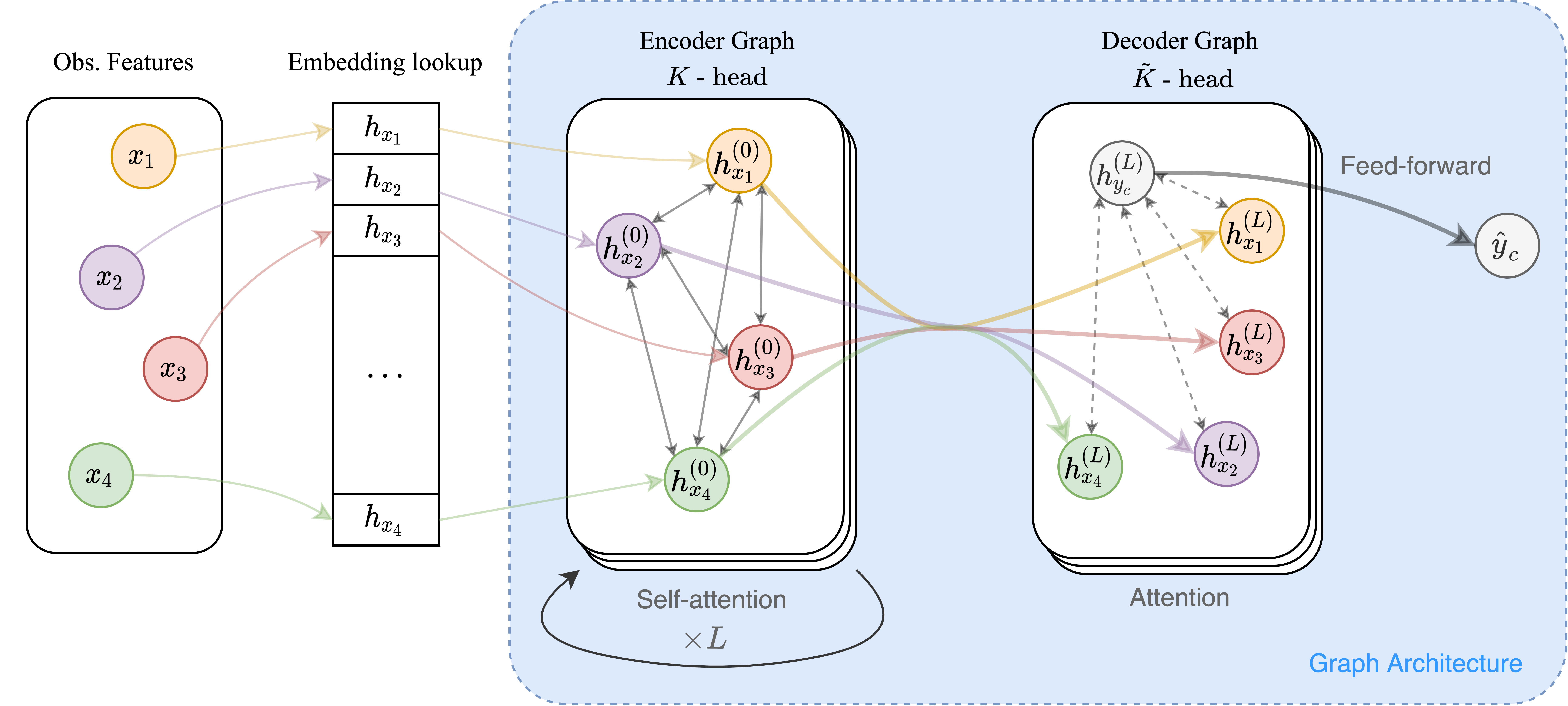}}
  \subcaptionbox{The illustration of variational regularization (\textit{VGNN}) on the graph architecture (blue highlighted block in (a))}{
  \includegraphics[width = 0.7\textwidth]{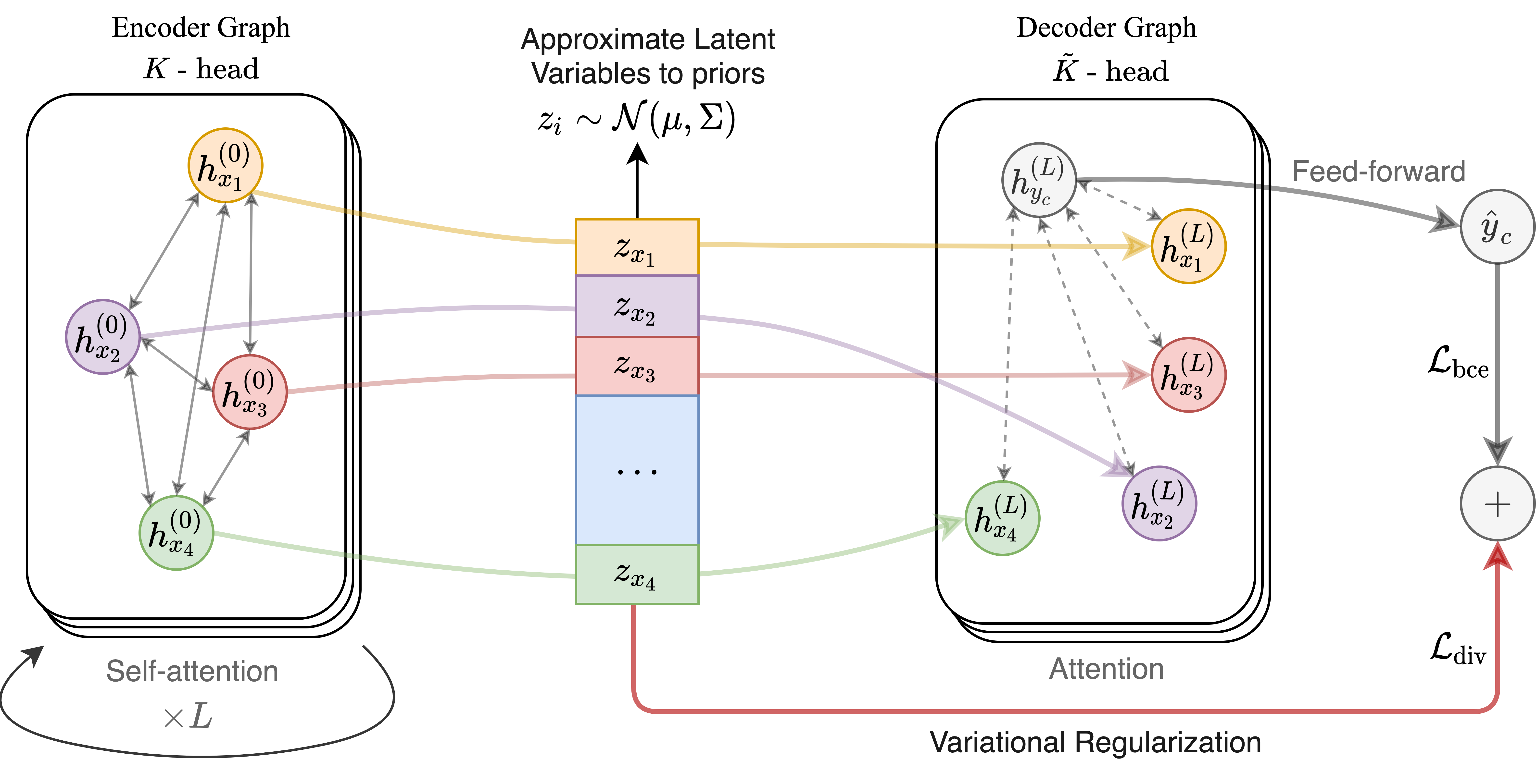}}
  \caption{The architecture of our encoder-decoder model (top architecture), as well as variational regularization (bottom architecture). For each patient, the observed features $i \in \mathcal{V}_{\text{obs}}$ ($\{x_1,x_2,x_3,x_4\}$ in the illustration) correspond to the EHR variables that were documented/observed in the patient records. These features are encoded as embeddings $\{h_{i}\}_{i \in \mathcal{V}_{\text{obs}}}$. Note that each patient has a different $\mathcal{V}_{\text{obs}}$. These observed nodes are fully connected in the first layer. In subsequent layers, the
  weights on the edges $\{A_{ij}\}_{i,j \in \mathcal{V}_{\text{obs}}}$, and the graph representation $\{h^{(l)}_{i}\}_{i \in \mathcal{V}_{\text{obs}}}$, are computed by multi-head attentions. The "encoder graph" denotes the observed nodes and the parametrized attention-based connections up to layer L. In the decoder graph, we add new nodes corresponding to the target outcomes to the encoder graph's output. Inferences on the outcomes $\{\hat{y}_c\}_{y_c\in \mathcal{V}_{\text{{out}}}}$ are derives from multi-head attentions on node representation and a linear feed-forward layer.
  In variationally regularized model, a latent layer $\{z_i\}_{i\in \mathcal{V}_{\text{obs}}}$ is placed between the encoder and decoder graph. Latent variables $z_i$'s are sampled from $\mathcal{N}\left(\mu_i, \exp(\sigma_i)\right)$, where $\mu_i$ and $\sigma_i$ are computed from $h_i^{(L)}$ by two separate feed-forward networks. The latent variables are regularized by $\mathcal{L}_{\text{div}}$, approximating the distributions of $z_i$'s to some priors $\mathcal{N}(\mu, \Sigma)$ (we use standard Gaussian here).}
  \label{model_structure}
\end{figure*}
\subsection{Encoder-decoder Graph Neural Networks}
We embed EHR codes $\mathcal{V} = \{1,2, \cdots, N\}$ with high-dimensional vectors $\{h_i\}_{i\in \mathcal{V}}, h_i \in \mathbb{R}^{d}$. Compared to GAT
\cite{velickovic2018graph} which takes advantage of external features of nodes, we learn the representations of each medical concept. For each patient $X$, we denote their observed codes $\mathcal{V}_{\text{obs}} = \{x_1, x_2, \cdots, x_n\}$ and initially fully connect them, and consequently learn the structure and updated representations over $L$ additional graph layers. This part of our architecture is termed the encoder graph. We then denote additional nodes $\mathcal{V}_{\text{{out}}} = \{y_1, y_2, \cdots, y_m\}$ for prediction task and fully connect them to output nodes of encoder graph. This sub-network is termed the decoder graph. As outlined in the top architecture of Figure \ref{model_structure}, the medical embeddings are processed by the encoder to represent the graph, and then the decoder provides inferences based on the graph representations. In each graph layer, the representations are updated by graph propagation.
\begin{equation} 
\label{graph_convolution}
H^{(l+1)} = \text{FFN}\left[A^{(l)}\left(H^{(l)}W^{(l)} + b^{(l)}\right)\right]
\end{equation}
where $W^{(l)} \in \mathbb{R} ^ {d\times d}$ and $b^{(l)} \in \mathbb{R} ^ {d}$ form a linear layer; $H^{(l)}$ is the matrix taking all the representations $h_i^{(l)}$ of observed nodes as row vectors; $A^{(l)}$ is the adjacency matrix at the $l^{\text{th}}$ layer. The size of $H^{(l)}$ and $A^{(l)}$ varies with the sample and location of the layer. The nodes in the graph are $\mathcal{V}_{\text{obs}}$ in the encoder and $\mathcal{V}_{\text{obs}} \bigcup \mathcal{V}_{\text{{out}}}$ in the decoder. Hence, $A^{(l)} \in \mathbb{R}^{n\times n}, H^{(l)} \in \mathbb{R}^{d\times n}$ in the encoder, and $A^{(l)} \in \mathbb{R}^{(n+m)\times (n+m)}, H^{(l)} \in \mathbb{R}^{d\times (n+m)}$ in the decoder. $FFN$s, referring to feed-forward networks are the multilayer perceptron composed of linear layers, ReLU activations, dropout \cite{JMLR:v15:srivastava14a} and layer normalization \cite{DBLP:journals/corr/BaKH16}. In any graph layer $l$, the elements of adjacency matrices $A_{ij}$ on the edge connecting node $i$ to node $j$ are computed using attention mechanism.  
\begin{equation} 
A_{ij} = \text{softmax}(e_{ij}) = \frac{\exp{(e_{ij})}}{\sum_{p \in \mathcal{N}_i}\exp{(e_{ip})}}
\label{attention}
\end{equation} 
where $e_{ip} \in \mathbb{R}$ are the attention coefficients for node $i$ over its neighbourhood $\mathcal{N}_i$. There are multiple ways of computing attention coefficients with interactions between two input vectors \cite{bahdanau2014neural, DBLP:journals/corr/LuongPM15, NIPS2017_7181}. The selection of attention style is discussed in Section \ref{discussion}. In this study, we use multi-head attention, as follows: We first concatenate two input vectors and apply a linear layer $a\in \mathbb{R} ^ {2d \times 1}$. The attention coefficients are computed by the following equation:
\begin{equation} 
e_{ij} = \text{LeakyReLU}\left(a^T\left[Wh_i \mathbin\Vert Wh_j\right]\right)/\sqrt{d_h}
\label{alignment}
\end{equation}
$W$ is a linear layer, and $d_h$ is the dimensionaility of $h_i$'s.

Multi-head attention \cite{NIPS2017_7181} allows the model to jointly attend to information from different representation subspaces at different positions like multiple convolution filters in one CNN layer. To build our $K$-head attention, we alter the output size of linear layer $W^{(l)}$ and $b^{(l)}$ in equation (\ref{alignment}) to $dK$, and then attention heads can be computed in parallel. Also, the outputs of multihead attention are aggregated by concatenation, so the input size of the feedforward networks is adjusted to $dK$ as well to fit the concatenated representations.

As the majority of predictive tasks in EHR have imbalance labels (more negative than positive), we use the weighted binary cross-entropy loss to train this model:
\begin{equation}
    \label{eq:bce}
    \mathcal{L}_{\text{bce}} = -\sum_{y_c \in \mathcal{V}_{\text{{out}}}} w_c \cdot Y_c \cdot \text{log} \left [ \sigma(\hat{y}_c )\right]+(1-Y_c) \cdot \text{log}\left[1- \sigma (\hat{y}_c)\right]
\end{equation}
where $\sigma$ is the sigmoid function; $Y_c$ is the ground truth of $y_c$; $\hat{y}_c \in \mathbb{R}$ is the output of the last feed-forward layer in the decoder; $w_c$ is the negative-to-positive ratio, putting more weight on the minority. This loss function describes the loss of all the outcomes for one sample, and the loss of mini-batch is the mean of losses over the batch.
\subsection{Variationally regularized Encoder-Decoder Model}
In our experiments of the encoder-decoder graph networks, we observe that node representations after the encoder layer are often collapse to a tight clustered and lack implicit structures, which leads to uniformly-distributed attention weights and prevents graph layers from learning meaningful edges among medical concepts. The uniformity of the attention weights is also observed by Choi et al. \cite{DBLP:journals/corr/abs-1906-04716}, and solved by regularizing the links with a pre-defined knowledge graph. In this study, we trace the problem to the distribution of embeddings and introduce variational regularization to encourage the node representations to be centered around the origin with moderate distances, which as we will show, leads to learning more expressive connections.

Inspired by VGAE \cite{kipf2016variational} which improves link inference by assuming a Gaussian prior on the node representations, we add a latent layer between the encoder and decoder to regularize the graph representation. Let $Z = \{z_i\}_{i \in \mathcal{V}_{\text{\text{obs}}}}, z_i \in \mathbb{R}^d$ be the latent variables corresponding to each observed node representations after encoder layers $h^{(L)}_i$.  We assume a standard normal prior distributions $p(z_i) \sim \mathcal{N}(0, I)$ and the generative encoder distribution of $q(z_i|X) \sim \mathcal{N}(\mu_i, \exp({\sigma_i}))$ where $\mu_i$ and $\sigma_i$ are learned from encoder outputs with a linear layer (i.e. $\mu_i = W_{\mu} h_i^{(L)}+b_\mu$ and $\sigma_i = W_{\sigma} h^{(L)}_i+b_\sigma$). The variance is parameterized as an exponential to assure non-negativity. Then the sampled latent variables $z_i$'s, replacing $h^{(L)}_i$'s, become the inputs to the decoder layer. Let $p(Z)=\prod_{i\in Z}p(z_i)$ and $q(Z|X)=\prod_{i\in Z}p(z_i|X)$. The variational formulation on auto-encoders solves the maximization problem on the posterior $p(X|Z)$ by maximizing the Evidence lower bound (ELBO) \cite{kingma2013autoencoding}:
\begin{equation}
\label{eq:kl}
ELBO_{\text{VAE}}  = \mathbb{E}_q\left[\log p(\hat{X}|Z)\right] - \mathrm{KL}\left[ q(Z|X) \| p(Z) \right]
\end{equation}
 where the first term is the loss for reconstructing the input and $\mathrm{KL}(\cdot \| \cdot)$ is Kullback–Leibler divergence $\mathcal{L}_{\text{div}}$ between prior distribution and likelihood of the latent space $Z$. From an empirical perspective, the KL-term $\mathcal{L}_{\text{div}}$ regularizes $z_i$'s to center around the origin, while the reconstruction term ensures sufficient distance between the $z_i$'s to prevent mode collapse and retain expressiveness. Here, we use the KL-term $\mathcal{L}_{\text{div}}$ to regularize representations of medical concepts in our supervised model, and combine that with cross-entropy loss $\mathcal{L}_{\text{bce}}$ in Equation (\ref{eq:bce}) as the loss function to minimize:
 \begin{equation}
     \label{eq:vgnn_loss}
     \mathcal{L}(y,\hat{y}) = \mathcal{L}_{\text{bce}}(y,\hat{y}) +  \mathrm{KL}\left[ q(Z|X) \| p(Z) \right]
 \end{equation}
For the rest of the paper, we denote our encoder-decoder graph neural network as \textit{Enc-dec} and the variationally regularized formation as \textit{VGNN}. Our implementation is open source and available at: \url{https://github.com/NYUMedML/GNN_for_EHR}.
\section{Experiments}
In this study, we test the proposed methods in the context of three clinical tasks: readmission prediction at discharge, based on eICU cohort \cite{eicu}, mortality prediction at 24 hour after admission, based on MIMIC-III cohort\cite{mimiciii} and Alzheimer's Disease prediction within 12 to 24 months based on inpatient and outpatient EHR data from NYU Langone Health, (AD-EHR for short). The first two datasets consist of short-term records from ICUs (inpatient); the third dataset is a long-term inpatient and outpatient clinical EHR spanning over 4 years. With the experiments on various type of EHR data, we demonstrate the capacity of our method on EHR representation learning. 

All the EHR dataset are partitioned into training, validation and test set by patient unique IDs at ratio of 8-1-1, respectively. We train models based on training sets, tune hyperparameters with validation sets and report the performance of models with test sets.

\begin{table}[h]
  \caption{Dataset Statistics Summary (number of average observed features / number of total features)}
  \label{data_stats}
  \centering
  \begin{tabular}{llll}
    \toprule
    Dataset  & \textbf{AD-EHR} & \textbf{MIMIC-III} & \textbf{eICU}\\
    \midrule
    Diagnosis  &  10.1/6028 & 11.5/6778 & 6.5/3250\\
    Procedures & --- / --- & 4.5/2006 & 5.0/2212\\
    Lab Values & 6.1/3073 & 62.2/3032 & --- / ---\\
    Demographic & 3.3/38 & --- / --- & --- / --- \\
    \midrule
    
    $\#$ of positives  & 8174 & 5377 & 7051\\
    $\#$ of total patients & 1613088 & 50391 & 41026\\
    \bottomrule
  \end{tabular}
\end{table}
\begin{table*}
  \caption{Model evaluation on the test set using precision-recall curves (99\% confidence interval)}
  \label{result}
  \centering
  \begin{tabular}{lcccc}
\toprule
  \multirow{2}{*}{\textbf{Method}} & \multicolumn{2}{c}{\textbf{AD-EHR}} & \textbf{MIMIC-III Mortality} & \textbf{eICU Readmission} \\
      & AUPRC & PPV@0.4Recall & AUPRC & AUPRC \\
    \midrule
    Random Forest \cite{breiman2001random} & $0.2316\pm0.0043$ & $0.0890\pm0.0029$ & $0.5976\pm0.0056$ & $0.3614\pm0.0049$\\
    MLP\cite{mlp} & $0.3775\pm0.0050$ & $0.5623\pm0.0182$ & $0.6646\pm0.0045$ & $0.3639\pm0.0045$\\
    RNN* \cite{Diag_LSTM} & $0.2590\pm0.0045$ & $0.3038\pm0.0041$ & --- & ---\\
    CNN* \cite{razavian2015temporal} & $0.3566\pm 0.0053$ & $0.4267\pm 0.0056$ & --- & ---\\
    NBOW \cite{iyyer-etal-2015-deep} & $0.3386\pm0.0049$ & $0.5265\pm0.0138$ & $0.6787\pm0.0054$ &  $0.3730\pm0.0049$ \\
    Transformer \cite{DBLP:journals/corr/abs-1906-04716} & $0.3957\pm0.0044$ & $0.6844\pm0.0165$ & $0.6777\pm0.0051$ &  $0.3792\pm0.0042$ \\
    GCT \cite{DBLP:journals/corr/abs-1906-04716} & $0.3409\pm0.0040$ & $0.5174\pm0.0095$ & $0.6810\pm0.0046$ & $0.3794\pm0.0045$  \\
    \midrule
    \textbf{Enc-dec} (Ours) &  $0.4216\pm0.0047$ & $0.6756\pm0.0109$ & $0.6962\pm0.0051$ &  $0.3881\pm0.0047$\\
    \textbf{VGNN} (Ours)& $\pmb{0.4580\pm0.0048}$ & $\pmb{0.7489\pm0.0075}$ & $\pmb{0.7102\pm0.0046}$&$\pmb{0.3986\pm0.0050}$\\
    \bottomrule
  \end{tabular}
\end{table*}
\subsection{Alzheimer's Disease Prediction}
Alzheimer's Disease (AD) leads to the majority of dementia, but the cause of AD is poorly understood. This gap in disease mechanism prevents researchers from constructing a medical knowledge graph that associates AD-related diseases and variables. This motivates us to attempt to learn the graph connections from scratch. In this experiment, we use the EHR from NYU Langone Health corresponding to 1.64M distinct patients with unique Medical Record Numbers (MRN), spanning from 2016 to 2019. AD-EHR includes diagnosis, recorded as ICD-10 codes, and lab values recorded as LOINC codes. More details on cohort selection and data preprocessing are described in Appendix \ref{data}. After the data prepossessing, the whole dataset includes 1.61M patient records and the encounter-based records for each patient are transformed into a 9139-dimensional one-hot vector. The detailed statistics on feature distributions are presented at Table \ref{data_stats}. To assess whether aggregation across time lead to any major loss of information, we compare our models built on aggregated EHR data with two encounter-based time series baseline models. As presented in the results section, we find that for the AD-EHR task, aggregated EHR is more effective than time series data.
\subsection{MIMIC-III and eICU Predictive Tasks}
MIMIC-III and eICU data are two publicly available EHR datasets collected from ICU patients. There has been several clinically meaningful predictive tasks studied for these population, including mortality prediction, readmission prediction and phenotype classification. Choi et at. \cite{DBLP:journals/corr/abs-1906-04716} leverages graph structures of EHR in predicting readmission and mortality of ICU patients, using eICU data. We use the same prepossessing steps on eICU to compare our models with the published pre-defined guidance knowledge graphs. 
Besides eICU, we also empirically evaluate our methods on a more common public dataset, MIMIC-III. Mortality prediction at early days after ICU admission (i.e. 24 hours or 48 hours) is among widely studied and clinically useful predictive tasks based on MIMIC-III dataset \cite{mimic1, mimic2, DBLP:journals/corr/ChePCSL16}, although the selection on feature sets varies a lot in different studies. We follow the schemas used in AD-EHR and eICU and make the features more cohesive. We not only extract ICD-9 codes and CPT procedure codes referring to the schema of eICU, but also categorize lab values into buckets according to the schema of AD-EHR. To avoid potential data leakage between mortality and the preventative events immediately preceding it, we only include the chart events within the first 24 hours after ICU admission as the input for the mortality prediction task. This clinical task setting is based on the benchmark study by \cite{mimic2}. 
\subsection{Baseline Models}
We introduce several baseline models in various machine learning domains to demonstrate the necessity of our design on the model architecture and the statistically significant improved performance of our methodology.
\begin{itemize}
    \item \textbf{Random Forest} Random forest \cite{breiman2001random} is an ensemble model of decision trees that takes advantage of bagging mechanism to reduce overfitting. It exams whether deep learning methods are over-complex.
    \item \textbf{Multilayer Perceptron}
    MLP is the multiple feed-forward network, previously used in predictive tasks in EHR \cite{mlp}.
    We use MLP as a non-embedding baseline that takes one-hot vectors of disease codes as inputs.
    \item \textbf{Temporal Methods*} 
    We use two temporal deep learning methods on AD-EHR to investigate the impact of time: an RNN with 3-layer LSTM \cite{Diag_LSTM}, and a CNN with convolutional block \cite{razavian2015temporal} in Table \ref{tab:cnn} (Appendix) on both feature axis and time axis.
    \item \textbf{Neural Bag of Words} 
    Using the average or the sum of embeddings \cite{iyyer-etal-2015-deep,Mikolov2013EfficientEO} is a method for representing patients.
    We embed the medical concepts with embeddings $h_i$, and represent a patient by averaging embeddings of all the positive features \cite{iyyer-etal-2015-deep,Mikolov2013EfficientEO}. Then a feed-forward network outputs the inference on target classes.
    \item \textbf{Transformer}
    Choi et al. \cite{DBLP:journals/corr/abs-1906-04716} adapt Transformer \cite{NIPS2017_7181} to learn the graph structures of EHR via interactions among the medical concept embeddings $h_i$'s and the visit embeddings $v$'s. This work also shows that Transformer has superior performances in EHR predictive tasks than graph convolution networks (GCN) with pre-defined graphs and random graphs.
    \item \textbf{Graph Convolutional Transformer} GCT \cite{DBLP:journals/corr/abs-1906-04716} takes advantage of pre-defined medical ontologies as the prior guidance to regularize the attention weights in Transformer. The prior graphs takes the medical concepts as vertices and the connections between diagnosis and procedures, procedures and lab values as edges. The edges are weighted by empirical conditional probabilities derived from the co-occurrence relationship among medical concepts. We use the publicly available codes and hyperparameters to train this baseline.
\end{itemize}
\begin{figure*}[!b]
    \centering
    \includegraphics[width=\textwidth]{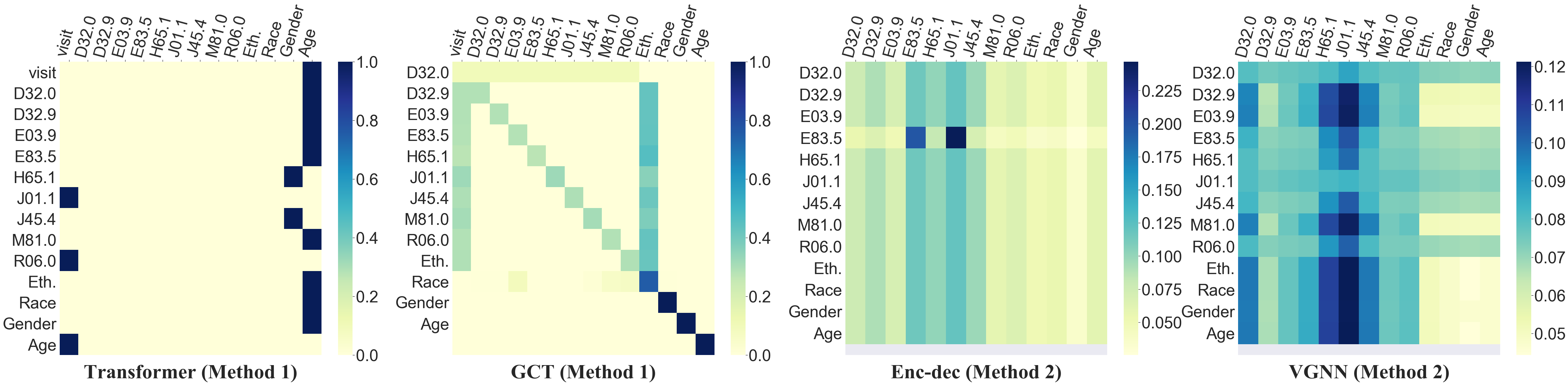}
    \caption{The patterns of the first layer graph with different attention functions. Method 1 is the attention used in Transformer; method 2 is the attention in our models. Results are visualized based on trained model representations of a randomly selected patient from the held-out test set of the AD-EHR task. The patient has between 10 to 20 observed features and a positive outcome label.}
    \label{fig:attention}
\end{figure*}
\begin{figure}[!t]
    \centering
    \includegraphics[width=0.47\textwidth]{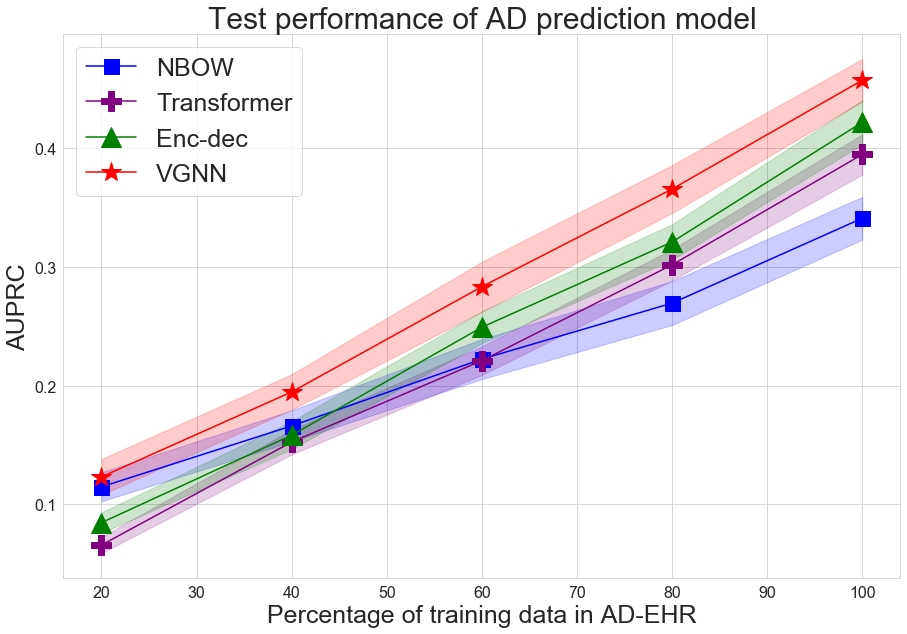}
    \caption{The evaluation of graph and non-graph based methods on different levels of training data sizes using AD-EHR. The shaded area denotes $\pm$ one standard deviation.}
    \label{fig:data_size}
\end{figure}
\begin{figure*}[!b]
    \centering
    \includegraphics[width=0.85\textwidth]{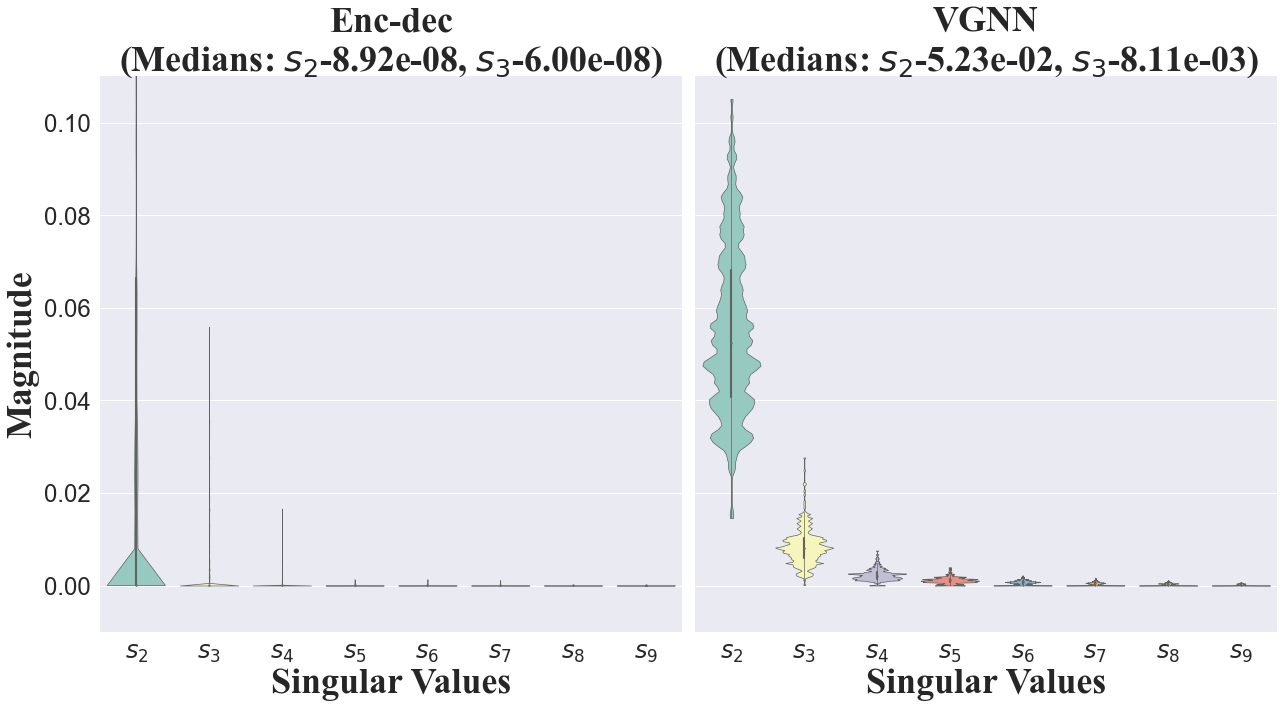}
    \caption{The magnitudes of singular values of the first graph convolution layer on AD-EHR data. The black vertical lines show the range, and the curves in the vertical direction show the smoothed distribution of magnitudes. Note that more than half of the first singular values of Enc-dec vanish.}
    \label{fig:singluar-values}
\end{figure*}
\begin{figure*}[!t]
    \centering
    \subcaptionbox{Patient with less significant singular values}{
    \includegraphics[width=0.45\textwidth]{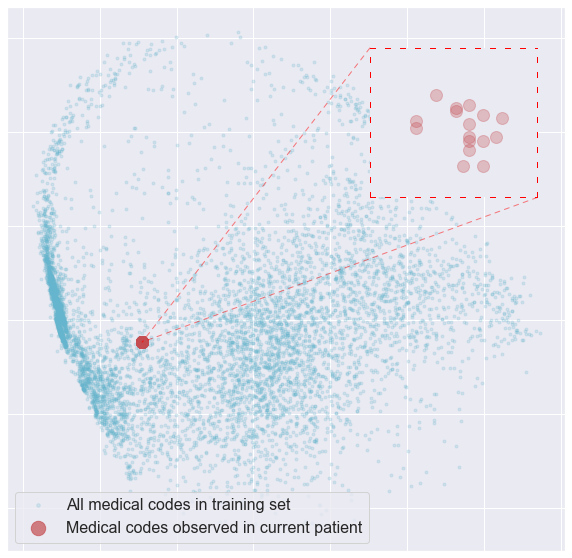}}
    \subcaptionbox{Patient with more significant singular values}{
    \includegraphics[width=0.45\textwidth]{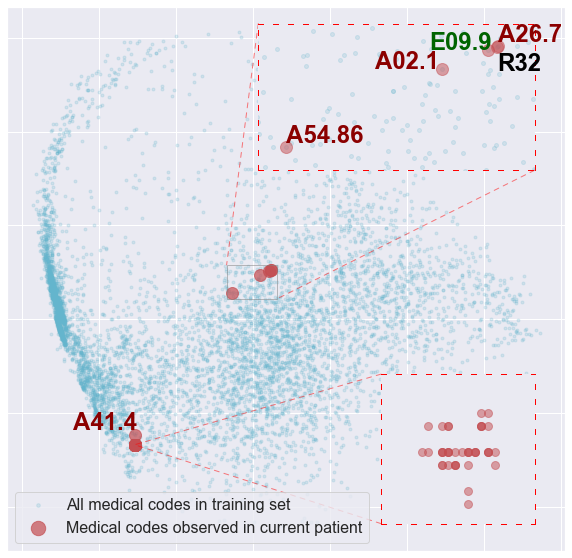}
    }
    \caption{
    \textit{2D} PCA projection of learned representations of $h_{i}$ after graph encoder layer in the VGNN model. Blue dots represent projections of every feature (medical code) observed in the training cohort. In red, we overlay the observed features of two different patients, one with a low count of non-zero singular values (left), and one with high count of non-zero singular values (right). We observe that the increment in non-zero singular values corresponds to more clusters in the projection of the learned representation of the patient features.
    The clusters that form for each patient also exhibit meaningful medical meanings: i.e. The red nodes with annotations in the right plot correspond to to sepsis (red annotated ICD codes), diabetes (green annotated ICD codes) and urinary incontinence(black annotated ICD codes).}
    \label{fig:cluster}
\end{figure*}
\subsection{Results}
In Table \ref{result}, we report the performance of different models on three tasks. Since all of three tasks have imbalances class labels, the precision-recall curve is a more informative evaluation metric on the prediction performance than ROC curve \cite{10.1371/journal.pone.0118432}. To quantify PR-curve, we compute the area under PR-curve (AUPRC) to summarize the curve. We compute the mean and confidence interval by bootstrapping the test data 100 times. The results in Table \ref{result} show that the 99\% confidence intervals of our VGNN method have no overlap with the baseline models, indicating VGNN outperforms the other baseline models statistically significantly. The optimal hyperparameter settings are in Table \ref{tab:hyperparameters} (Appendix), and the precision recall curves for three tasks are shown in Figure \ref{fig:pr-curve} (Appendix). We observe that the precision for AD-EHR task have a sharp drop around 0.4 recall, so we introduce PPV@0.4Recall to depict the precision at a relatively high classifier threshold for AD-EHR. 
Table \ref{result} shows the graph-based models are superior to the simple embedding model like NBOW. This comparison demonstrates the importance of connections among different medical concepts. We notice that for the AD-EHR prediction, the performance of GCT are worse than other graph-based methods and close to NBOW. This problem is cause by the nature of dataset: unlike the data from ICU where the variables are measured frequently, AD-EHR is primarily outpatient clinical data. It has more lab values missing and most patients only have diagnosis codes. However, the design of GCT prunes the connections among diagnosis codes, so when lab measures are missing, CGT will reduce to NBOW on diagnosis codes. The insufficiency of GCT on AD prediction demonstrates the importance of learning connections among diagnosis codes. These connections can be overlooked in short-term ICU records, but they are in fact crucial in learning the overall graph representation of patients. The interaction among the code representation helps reduce the impact of missing codes. 
The lower performance of the temporal models on AD-EHR shows that for long-term EHR tasks, the temporal information is not a dominating signal.

In addition to the comparison across methods on the same task vertically, we also compare and analyze the results among different tasks. The size of public EHR datasets, like eICU and MIMIC-III, are limited, while in the real-world the hospital system usually has more patients in the EHR database. Table \ref{result} shows that not all of the graph based methods dominate other non-graph based methods such as bag-of-words, while the experiments on AD-EHR demonstrates outstanding performances of models that learn appropriate graphs. The size of dataset can be a cause on this phenomenon. Hence, we also evaluate the model capacity with various data sizes. AD-EHR includes 1.6M patients records, which allows us to analyze the impact of the size of dataset on learning graph structures, by training models with different sizes of training data. We experiment training with different proportion of AD-EHR and evaluate the performance change on the test set. Figure \ref{fig:data_size} shows that at the low data size level, NBOW is only slightly inferior to VGNN. However, as the training data size grows, the performance of VGNN far exceeds NBOW. Also, the slope of curves in Figure \ref{fig:data_size} indicates that the graph-based models have more performance gain with the increment on data size than NBOW. This finding explains why graph-based methods have the most performance gain in AD prediction among all three tasks. It also indicates that learning the medical graph connections enlarges the model capacity, so the graph based model has more potential to improve by learning from more data. Therefore, even though the improvement of incorporating graph structure on MIMIC-III and eICU are less than AD-EHR, it follows the trend of performance growth with the size of data.
\section{Discussion}
\label{discussion}
In this section, we discuss the main components of the development of our models that improve the model performance and provide support for our choices of various settings. We develop the analysis on the methods with both quantitative statistics and qualitative interpretation. 
\subsection{Behavior of Attention Functions}
Attention mechanism is widely used in deep learning to express the ``soft" links among representations. It is a collections of functions $f: \mathbb{R}^d \times \mathbb{R}^d \longrightarrow [0,1]$. There are several common ways to compute attentions in previous literature. Transformers model \cite{NIPS2017_7181} uses feed-forward networks to create three vectors - key $K$, query $Q$ and value $V$. The attention weights are computed by (Method 1):
\begin{equation}
    \label{transformer}
    \text{Attention}(Q,K,V) = \text{softmax}\left(\frac{QK^T}{\sqrt{d_k}}\right)V
\end{equation}
Our method in Equation (\ref{alignment}) is in another attention style \cite{bahdanau2014neural, DBLP:journals/corr/LuongPM15, velickovic2018graph} (Method 2) that takes the inner product of a learnable vector and the concatenation of two relevant representations. Table \ref{result} shows the performance difference between two methods, Transformer/GCT (Method 1) vs. the Enc-dec/VGNN (Method 2). We unveil the functional properties of two attention functions by analyzing the graph structure patterns of two attention mechanisms. Figure \ref{fig:attention} shows that attention function in Equation (\ref{transformer}) directs the model to a sparser graph that only includes 0 and 1. This pattern is desirable for tasks such as alignment in machine translation where there are one-to-one bijections between words most of the times. However, in the depicted EHR example, the ``hard" edges lead to a problem that the 14 node representations in the example are reduced to 3 node representations.

The attention mechanism in our model (Method 2) is more robust: in the sample of VGNN, we can observe the that head-related diseases D32.0 (Benign neoplasm of cerebral meninges), H65.1 (Other acute non-suppurative otitis media) and J01.1 (Acute frontal sinusitis) have more attention weights, while the representations of the other nodes are also included in the outputs towards deeper layers. In general, compared to method 1, the attention matrix computed by method 2 has positive weights on multiple medical concepts, so the graph representations are able to receive the message different medical embeddings. In Figure \ref{fig:attention}, we also notice that some row attention weights in Enc-dec are numerically close to each other. This phenomenon leads to the graph representation to be similar. We then quantitatively analyze the impact of this phenomenon and how the problem is solved by the regularization of VGNN in Section \ref{singular-value}.
\subsection{Singular Value Analysis on Graphs}\label{singular-value}
We now assess the performance of the attention mechanism quantitatively by decomposing the graph attention layer. The adjacency matrices $A^{(l)}$ contain the graph structural information learned by the GNN. Previous studies show that the spectral analysis of graph convolution kernels can lead to a deeper understanding of the Laplacian matrices in graph convolution \cite{chen2020bridging, balcilar2020bridging}. Similarly, to characterize the learned $A^{(l)}$'s, we start by using singular value decomposition (SVD):
\begin{equation}
    A^{(l)} = U \text{diag}(s_i)_{i \in \mathcal{V}_{\text{obs}}}V^T
\end{equation}
where $U, V$ are the collections of orthonormal vectors and $s_i$ are singular values of $A^{(l)}$. Through SVD, the graph convolution can be transformed to a linear combination of the node representations' projections onto subspaces spanned by these orthonormal basis. 
\begin{equation}
    A^{(l)} H^{(l-1)}_{:,j} = \sum_{i = 1}^{|\mathcal{V}_{\text{obs}}|} s_i u_i v_i^T H^{(l-1)}_{:,j}, \left (1 \le j \le d \right)
    \label{svd_expand}
\end{equation}
where $H_{:, j}$ are the collections of the $j^{th}$ elements in node representations at the layer $l-1$. The magnitudes and directions of transformation given by graph attention layers can be interpreted from singular values and orthonormal basis $U$. Hence, from Equation (\ref{svd_expand}) we learn that the singular values of $A^{(l)}$ correspond to the magnitude of the messages passed among nodes in the graph layer. 

The softmax normalization on attention coefficients restricts adjacency matrices $A^{(l)}$ such that $\sum_{j\in \mathcal{V}_{\text{obs}}}A^{(l)}_{ij} = 1$. Under this constrain, we can have a lower bound on the largest singular value.
\begin{lemma}
\label{lemma}
Let matrix $A \in \mathbb{R}^{d\times d}$. Suppose $\sum_{j=1}^d A_{ij} = 1, 1\le i\le d$, and $A$ has singular values $s_1 \ge s_2 \ge \cdots \ge s_d$, then $ s_1 \ge 1$.
\end{lemma}
\begin{proof}
Appendix \ref{lemma-proof}.
\end{proof}
By Lemma \ref{lemma}, all the first singular values of graph adjacency matrices are greater or equal than 1. Hence, we analyze the distribution of the remaining singular values to interpret the adjacency matrices, as follows: We perform SVD on the learned adjacency matrices of the heldout testset patients in AD-EHR dataset. We limit this analysis to patients with a positive label, who had more than 10 observed features. In Figure \ref{fig:singluar-values}, the singular values of first graph layers are visualized. The plots demonstrate that singular values of Enc-dec model has a larger range, but the majority of eigenvalues are close to 0. Vanishing singular values leads to fewer effective dimensions in the graph layers, according to Equation
(\ref{svd_expand}). As depicted in Figure \ref{fig:singluar-values}, more than half of the samples only have one non-zero dimension under Enc-dec. But we observe that they contain at least two non-zero singular values using VGNN. Also, in VGNN model, we can observe that most analysed patients have at least 5 singular values significantly greater than 0. It indicates that the variational regularization enables the graph kernel to avoid mode collapse, and combine more node representations during inference, which helps with generalization of the model across different samples.

Figure \ref{fig:cluster} shows different clustering behavior of the learned node representations, at varying numbers of non-zero singular values. We visualize the 2D PCA projection of learned $h_{i}$ representations after graph encoder layer in the VGNN model. Blue dots represent projections of every feature (medical code) observed in the training cohort. In red, we overlay the observed features of two different patients, one with a low count of non-zero singular values (left), and one with high count of non-zero singular values (right). We observe that the increment in non-zero singular values corresponds to more clusters in the projection of the learned representation of the patient features. The clusters that form for each patient also exhibit meaningful medical meanings: i.e. The red nodes with annotations in the right plot correspond to to sepsis (red ICD code annotations), diabetes (green ICD code annotations) and urinary incontinence (black ICD code annotations). Previous medical studies indicate that these diseases are related to AD \cite{septicemia, diabete} and are correlated \cite{urinary}. This example helps demonstrate that a graph with more significant singular values aids learning more expressive representations, and we can interpret the semantics of the nodes based on the emerged clusters. 
\subsection{Interpretation of Variationally Regularized Graph Representations}
According to variational regularization, and as seen in Equation (\ref{eq:vgnn_loss}), the learnable parametrized distribution of representations of each observed code is forced closer to a standard Gaussian prior. This strategy prevents the representations from converging to only one direction in high dimensional space. Looking at the patients to which we applied SVD in Figure \ref{fig:singluar-values}, we measure the compactness of clusters by the mean of $\ell_2$ distances each points to the mass center after the encoder graph layer. \begin{equation}
    \text{compactness} = \frac{1}{M}\sum_{i=1}^M\sum_{i\in\mathcal{V}_{\text{obs}}}\frac{\Vert{h^{(l)}_i - c(h_i^{(l) })}\Vert_2}{\vert\mathcal{V}_{\text{obs}}\vert}
\end{equation}
where $M$ is the number of samples, and $c(\cdot)$ is the mass center of all node representations at given layer.
The compactness of node representations in Enc-dec is 0.5786, while in VGNN the compactness is 3.1036.  Additionally, for the same samples, we visualize the distribution of the values of the learned node representations. Figure \ref{fig:embedding_distribution} in Appendix \ref{Precison-Recall Curves} shows that the representations in Enc-dec only learned by self-attention are biased. 

Together, these two statistics indicate that the KL-term regularizes the representations to avoid over-clustering and biases. 
\section{Conclusion}
Bridging connections among medical concepts contributes towards learning more expressive representation of the EHR, and therefore, improves the performance on predictive tasks in population health. We proposed an encoder-decoder graph neural network that adaptively learns the connections among observed medical codes in EHR. Our method also addresses the problem of learning more expressive representations via variational regularization. We showed that our model achieves superior performance on three EHR-based predictive tasks. Singular value analysis presented here helped explain some of the empirically observed benefits of our proposed regularization, compared to standard graph based methods. Our future studies include exploration of self-supervised learning to further improve generalization of graph based EHR representation learning.
\bibliographystyle{ACM-Reference-Format}
\bibliography{sample-base}
\appendix
\section{Data details} \label{data}
We preprocessed 1.6M EHR with 100K features, including diagnosis, labs and procedures. Since the ICD-10 codes are hierarchical and get granular at different levels, we merge the codes that share same number up to the first place after the decimal point to the first codes in their subdivisions. We set the target variable by aggregating all the ICD-10 codes for AD \footnote{Agency for Healthcare Research and Quality(AHRQ) at United States Department of Health and Human Services defines the family of Alzhiemer's related dementia, including ICD-10 codes: F01.50, F01.51, F02.80, F02.81, F03.90, F03.91, F04, F05, F07.0, F07.81, F07.89, F07.9, F09, F48.2, G30.0, G30.1, G30.8, G30.9, G31.01, 
G31.09, G31.1, G31.83, R41.81, R54.} and exclude all these codes from our feature set. The original data is encounter-wise. We track each patient by partitioning his/her encounters into history window (before 2016.02.19), feature window (2016.02.20 - 2017.02.19) and gap window (2017.02.20 - 2018.02.19). We use observations from encounters in the feature window as our inputs, and exclude patients who are AD positive within any three windows. These patients are dropped to avoid data leakage, as our goal is to predict new-onset AD 12-24 months in the future. We then aggregate all the encounters in the feature window temporally to allow the model to focus on learning graph representations between the features, rather than focusing on sparsity patterns in the time dimension.

We use the following schema to aggregate the encounter data. The diagnosis features are set to observed/positive if they have positive outcomes in any encounter. The lab values are binned into ranges of -10, -3, -1, -0.5, 0.5, 1, 3, 10 standard deviations where the statistics of each lab are computed independently from training set. The lab value features are defined as observed/positive if the lab values of any encounter fall into the corresponding range.

\section{Training details}\label{Training details}
Graph neural networks can be memory consuming. In our setting, as the graph is fully-connected, for $O(n)$ nodes, $O(n ^2)$ edge weights should be allocated, where $n$ is the number of positive/observed features. However, for each patient graph, only a few nodes have observed values, so we implement the model in sparse form with Pytorch 1.1.0 to free the memory of unobserved nodes and their edges in the graph of each patient.

To improve the robustness and ability of the model inferring missing features through graph, we randomly mask $10\%$ of nodes during training. Since all of our predictive tasks have imbalanced labels (Table \ref{result}), we use weighted cross-entropy loss based on class weights. For AD-EHR data, the labels are extremely imbalanced, so the weighted loss cannot effectively improve the performance. Hence, we upsample positive samples of the training set by 50 times to see more positive samples in given epochs. We also randomly downsample $80\%$ negative patients with age under 50 each epoch to accelerate training, as they may not contain significant signals related to AD. For MIMIC-III and eICU, since the label are less imbalanced, we only upsample the positive samples 2 time. Validation and test set retain their original distribution.

We tune the hyper-parameters including the number of heads (1-4) and layers of graph in graph based models (1-3), the number of layers in feed-forward networks (1-2), embedding sizes (128-1024), dropout rates $[0,1]$ and learning rates $[10^{-5}, 10^{-3}]$. The optimal hyper-parameters in our experiments are listed in Table \ref{tab:hyperparameters}. Learning rate decay is used to avoid overfitting. We half the learning rate if the AUPRC stops growing for two epochs. For these experiments, we use Tesla V100 GPUs to train our model.

\section{Proof of Lemma}
\label{lemma-proof}
\begin{lemma}
Let matrix $A \in \mathbb{R}^{d\times d}$. Suppose $\sum_{j=1}^d A_{ij} = 1$ and $A$ has singular values $s_1 \ge s_2 \ge \cdots \ge s_d$, then $ s_1 \ge 1$.
\end{lemma}
\begin{proof}
Let $e = (1,1,\cdots, 1)^T$, we have $A^Te = e$, because $\sum_{j=1}^d A_{ij} = 1$. Therefore, $1$ is an eigenvalue of $A^T$. Since $A$ and $A^T$ has the same eigenvalues with same multiplicities, $1$ is an eigenvector of $A$. \\
According to the Min-max theroem, let $\mathcal{X} \subseteq \mathbb{R}^d$,
$$s_1 = \min_{\text{dim}(\mathcal{X}) = d} \max_{\Vert x\Vert_2 = 1, x \in \mathcal{X}}\Vert Ax\Vert_2 = \max_{\Vert x\Vert_2 = 1}\Vert Ax\Vert_2$$
Suppose $\lambda$ is the greatest eigenvalue of $A$ and $x$ be an eigenvector corresponding to  $\lambda$ such that $\Vert x \Vert_2 = 1$, we have $\Vert Ax\Vert_2 = |\lambda|$. Therefore, $s_1 \ge |\lambda| \ge 1$.
\end{proof}
\section{Supplementary Figures and Tables}\label{Precison-Recall Curves}

\begin{figure}[b]
    \centering
    \includegraphics[scale=0.25]{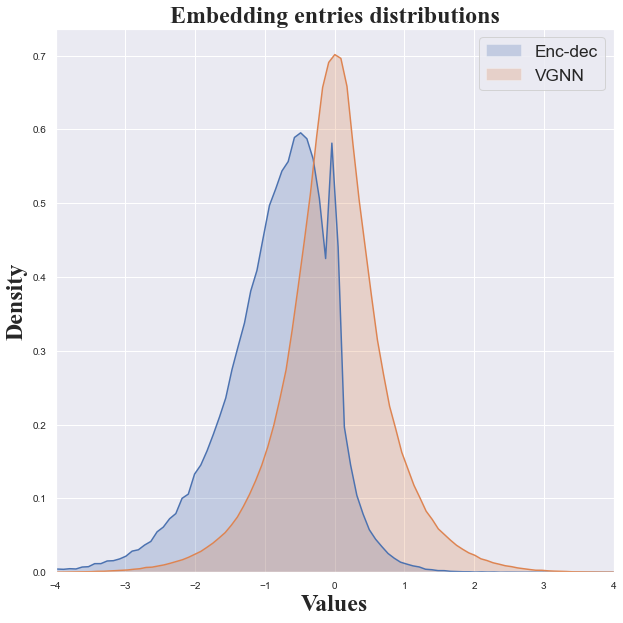}
    \caption{Distribution of node embedding entries in test samples of AD-EHR.}
    \label{fig:embedding_distribution}
\end{figure}
\begin{figure*}
    \centering
    \includegraphics[scale=0.4]{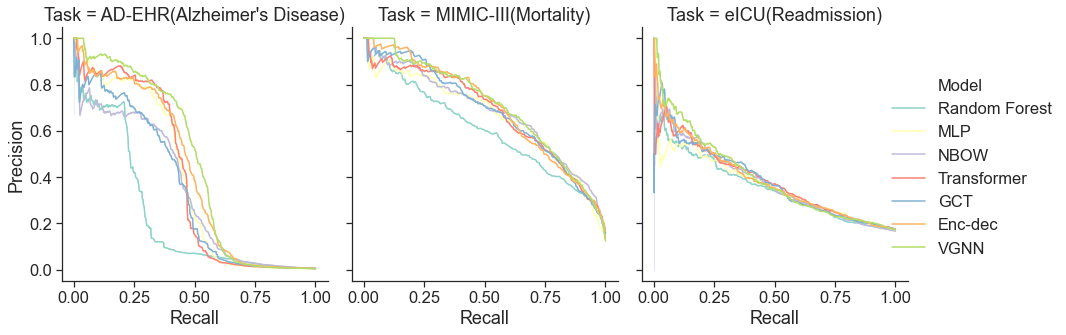}
    \caption{Precision-Recall curves of experiments corresponding to Table \ref{result}. The precision for AD-EHR task have a sharp drop around 0.4 recall, so we introduce PPV@0.4Recall to depict the precision at a relatively high classifier threshold for AD-EHR.}
    \label{fig:pr-curve}
\end{figure*}

\begin{table*}[b]
    \centering
    \begin{tabular}{llLLLLLLLLL}
        \toprule
        \textbf{Task} & \textbf{Model} & \textbf{No. Heads} & \textbf{No. Graph Layers} & \textbf{No. Feed-forward Layers} & \textbf{Learning Rate} & \textbf{Dropout Rate} & \textbf{Embedding Size} & \textbf{Batch Size}\\
        \midrule
         \multirow{5}{2cm}{\textbf{AD-EHR} \textit{Alzheimer's Disease Prediction}}  & MLP & --- & --- & 2 & 0.0001 & 0.3 & --- & 64\\
         & CNN* & --- & --- & 2 & 0.0003 & 0.2 & --- & 64\\
         & RNN* & --- & --- & 2 & 0.0003 & 0.2 & --- & 64\\
         & NBOW  &  --- & --- & 2 & 0.0003 & 0.3 & 1024 & 64\\
         & Transformer  &  1 & 3 & 2 & 0.0002 & 0.4 & 768 & 32\\
         & GCT  &  1 & 3 & 2 & 0.0002 & 0.1 & 768 & 32\\
         & Enc-dec  &  1 & 2 & 1 & 0.0001 & 0.4 & 768 & 32\\
         & VGNN & 1 & 2 & 1 & 0.00003 & 0.1 & 1024 & 32\\
        \midrule
         \multirow{5}{2cm}{\textbf{MIMIC-III} \textit{Mortality Prediction}}  & MLP & --- & --- & 2 & 0.0001 & 0.5 & --- & 64\\
         & NBOW  &  --- & --- & 3 & 0.0003 & 0.4 & 128 & 64\\
         & Transformer  &  1 & 3 & 2 & 0.0002 & 0.4 & 256 & 32\\
         & GCT  &  1 & 3 & 2 & 0.0002 & 0.1 & 256 & 32\\
         & Enc-dec  &  1 & 2 & 1 & 0.0001 & 0.4 & 768 & 32\\
         & VGNN & 1 & 2 & 1 & 0.0001 & 0.2 & 768 & 32\\
        \midrule
        \multirow{5}{2cm}{\textbf{eICU} \textit{Readmission Prediction}}  & MLP & --- & --- & 2 & 0.0001 & 0.5 & --- & 64\\
         & NBOW  &  --- & --- & 3 & 0.0003 & 0.4 & 128 & 64\\
         & Transformer  &  1 & 3 & 2 & 0.0002 & 0.45 & 128 & 32\\
         & GCT  &  1 & 3 & 2 & 0.00022 & 0.08 & 128 & 32\\
         & Enc-dec  &  1 & 2 & 1 & 0.0001 & 0.5 & 128 & 32\\
         & VGNN & 1 & 2 & 1 & 0.0001 & 0.4 & 128 & 32\\
        \bottomrule
    \end{tabular}
    \caption{Hyperparameters of experiments in Table \ref{result}. The hyperparameter settings of the previous study on the same dataset remain the same.}
    \label{tab:hyperparameters}
\end{table*}

\begin{table*}[h]
    \centering
    \begin{tabular}{llc}
        \toprule
    Type & Block & Hyperparameters \\
        \midrule
        \multirow{3}{1cm}{Vectical Conv} & Conv2d & k-$9139\times1$;c-64;p-0;s-1\\
        & ReLU & \\
        & BatchNorm2d & \\
        \midrule
        \multirow{3}{1cm}{Vectical
        Conv}& Conv2d & k-$64\times1$;c-128;p-0;s-1\\
        & ReLU & \\
        & BatchNorm2d\\
        \midrule
        Temporal& Maxpool1d & k-5;p-1;s-1\\
        \midrule
        \multirow{4}{1cm}{Temporal Conv}& Conv1d & k-5;c-256;p-1;s-1\\
         &ReLU\\
         &BatchNorm1d\\
         &Avgpool1d & k-T;p-1;s-1\\
        \midrule
        \multirow{4}{1cm}{FC$\times2$}&Linear & $256\times256$\\
        &ReLU \\
        &BatchNorm1d\\
        &Dropout\\
        \midrule
        &Linear & $256\times1$\\
        \bottomrule
    \end{tabular}
    \caption{Architecture of CNN for temporal signals. k is kernel size; c is the output channel; s is the stride; p is the padding size. T is the maximum number of encounters observed.}
    \label{tab:cnn}
\end{table*}
\end{document}